%% file: Main.tex
\documentclass[conference]{IEEEtran}
\IEEEoverridecommandlockouts
\usepackage{cite}
\usepackage{amsmath,amssymb,amsfonts}
\usepackage{algorithmic}
\usepackage{graphicx}
\usepackage{textcomp}
\usepackage{xcolor}

\usepackage{xcolor}
\usepackage{algorithmic}
\usepackage{amsmath,amssymb,amsfonts}
\usepackage[ruled,vlined,linesnumbered,noresetcount]{algorithm2e}
\input{math_commands.tex}

\newcommand{\fontstd}{\fontsize{7pt}{7pt}\selectfont} %
\newcommand{\fontstdd}{\fontsize{8pt}{8pt}\selectfont} %
\newcommand{\fontstddd}{\fontsize{8pt}{8pt}\selectfont} %
\newcommand{\fontsteqn}{\fontsize{9.5pt}{9.5pt}\selectfont} 

\def\BibTeX{{\rm B\kern-.05em{\sc i\kern-.025em b}\kern-.08em
    T\kern-.1667em\lower.7ex\hbox{E}\kern-.125emX}}

\begin{document}

\title{Diminishing Empirical Risk Minimization for Unsupervised Anomaly Detection}

\author{\IEEEauthorblockN{Shaoshen Wang$^{*}$, Yanbin Liu$^{\dagger}$, Ling Chen$^{*}$, and Chengqi Zhang$^{*}$}
\IEEEauthorblockA{\textit{}
$^{*}$Australian Artificial Intelligence Institute, University of Technology Sydney, Sydney, Australia\\
$^{\dagger}$Centre for Medical Research, The University of Western Australia, Perth, Australia\\
\{shaoshenwang, csyanbin\}@gmail.com, \{Ling.Chen, Chengqi.Zhang\}@uts.edu.au}
\thanks{
Corresponding author: Ling Chen.

This work was supported by ARC DP180100966 and DP210101347.
}
}

\maketitle

\begin{abstract}
Unsupervised anomaly detection (AD) is a challenging task in realistic applications. 
Recently, there is an increasing trend to detect anomalies with deep neural networks (DNN). However, most popular deep AD detectors cannot protect the network from learning contaminated information brought by anomalous data, resulting in unsatisfactory detection performance and overfitting issues. In this work, we identify one reason that hinders most existing DNN-based anomaly detection methods from performing is the wide adoption of the Empirical Risk Minimization (ERM). ERM assumes that the performance of an algorithm on an unknown distribution can be approximated by averaging losses on the known training set. 
This averaging scheme thus ignores the distinctions between normal and anomalous instances. To break through the limitations of ERM, we propose a novel Diminishing Empirical Risk Minimization (DERM) framework. 
Specifically, DERM adaptively adjusts the impact of individual losses through a well-devised aggregation strategy. 
Theoretically, our proposed DERM can directly modify the gradient contribution of each individual loss in the optimization process to suppress the influence of outliers, leading to a robust anomaly detector. 
Empirically, DERM outperformed the state-of-the-art on the unsupervised AD benchmark consisting of 18 datasets.
\end{abstract}


\section{Introduction}


Anomaly detection (AD) 
is an important research topic in data mining and machine learning \cite{ref:ADsurvey,ref:AD_survey2,ref:Deep-AD_survey}. It aims to identify data points that do not conform to expected behaviors. 
Since anomalies usually provide critical information, AD has been widely-used in various applications, such as health care~\cite{ref:health,ref:un-ad-medical}, network intrusion detection~\cite{ref:ad-network-intrusion,ref:un-ad-intrution}, fraud detection~\cite{ref:fraud} and other areas~\cite{ref:add1,ref:add2}.
In many realistic scenarios, there is no ground truth available to distinguish anomalous instances from the normal ones. 
The only assumption is that the proportion of normal instances is much higher than that of anomalies in a given dataset. 
According to this assumption, researchers usually resort to unsupervised approaches to cope with the situation, i.e. unsupervised anomaly detection (AD). 

A multitude of methods have been proposed to tackle the unsupervised AD problem ~\cite{ref:Deep-SVDD,ref:pca-ae,ref:dagmm,ref:un-ad-gan,ref:un-ad-time-series}. Recently, there is an increasing trend to use Deep Neural Networks (DNN) to solve the problem of unsupervised AD, as DNN-based methods show improved performance compared with traditional machine learning models, especially when the scale of data increases. For example, deep autoencoder has become the core of most deep unsupervised AD approaches~\cite{ref:Deep-AD_survey,ref:ADsurvey}, as it can powerfully extract and preserve intrinsic information from data. Specifically, an autoencoder learns the latent representation from the original data by minimizing the reconstruction loss. The anomalies are then detected based on the assumption that normal instances are more likely to be compressed and reconstructed than the anomalous ones.

Despite the progress, there  remains a major issue for deep AD detectors: most existing deep autoencoder AD methods cannot prevent the network from aggressively fitting the anomalies, which leads to the overfitting issue and the unsatisfactory performance in turn. One of the obvious reasons is that DNN are often over-parameterized and designed with non-linear activation function between layers, which makes DNN an universal approximator~\cite{ref:multilayer} fitting well with not only normal data but also the anomalies~\cite{ref:general}. Recent works~\cite{ref:RCA} have attempted to mitigate this issue by an elaborate design for model capacity. 

In this work, we analyze that another reason which makes DNN non-ideal for the scenario of unsupervised AD is the concept of empirical risk minimization (ERM), which is widely adopted by DNN. 
ERM assumes that the performance of a learning algorithm on an unknown data distribution can be approximated by averaging the losses on the known training set, as follows:
\begin{equation}
\label{eq:erm}
\bar{R}(\theta) := \frac{1}{N} \sum\limits_{i \in [N]} f(x_i;\theta)\,,
\end{equation}
where $f(\cdot;\theta)$ is a loss function parameterized by $\theta$ and $x_i$ is the $i$-th training instance from a dataset of size $N$. Take the deep autoencoder anomaly detector as an example. An anomalous instance $x_A$ tends to have a larger reconstruction loss $f(x_A;\theta)$. Under the ERM scheme, since the equivalent weights (i.e., $\frac{1}{N}$) treat all data equally during loss aggregation, an anomalous instance contributes even more to the overall loss $\bar{R}(\theta)$ than a normal instance. Consequently, the model parameters $\theta$ are optimized based on a total loss $\bar{R}(\theta)$ that is severely influenced by the losses incurred by  anomalies, making the deep model wrongly focus on and fit with the unwanted anomalies.
In other words, ERM ignores the distinctions between the normal and the anomalous instances.
While some methods~\cite{ref:RCA,ref:trimmed} have proposed to address this problem by either discarding or assigning low weights to potential anomalies during the training process, they are hard to well generalize on test data because these methods rely on inflexible ad-hoc or manual selection of potential anomalous instances.

To tackle the problem brought by ERM, we propose a novel Diminishing Empirical Risk Minimization (DERM) framework to adaptively adjust the impact of each individual loss. 
For $t \in \mathbb{R}^{+}$, DERM takes the following form:
\begin{equation}
\label{eq:DERM}
\textnormal{$ \tilde{R}_{\text{DERM}}(t;\theta) := $ }
e^{\frac{1}{N} \sum\limits_{i \in [N]} \log (t f(x_i; \theta))}. 
\end{equation}
The effectiveness of our design in Eq.~\ref{eq:DERM} comes from the intrinsic property of the logarithm function. 
Specifically, logarithm function is a slowly increasing function (i.e., its derivative is decreasing). 
When $f(x_i; \theta)$ has a larger value (usually for anomalies), the logarithm output $\log(t f(x_i;\theta))$ will suppress $f(x_i; \theta)$ more quickly. 
And the parameter $t$ controls how intensively the suppression will take effect. 
In this way, Eq.~\ref{eq:DERM} weakens the impact of potential anomalies (leading to larger loss values) in a dynamic and controllable manner. 
Moreover, by comparing the gradients of ERM (Eq.~\ref{eq:erm}) and DERM (Eq.~\ref{eq:DERM}) w.r.t $\theta$, we theoretically find that DERM can directly modify the gradient contribution of each individual loss term. 
In particular, if the loss $f(x_i;\theta)$ is larger than the average (i.e., $\tilde{R}_{\text{DERM}}(t;\theta)$), the gradient will be diminished. With this property, the potential anomalies will contribute less to the optimization process, leading to a more robust anomaly detection model. 

In the experiments, we first verify the effectiveness of the proposed DERM on anomaly suppression and gradient diminishing using both the synthetic and real-world datasets. And then we conduct comprehensive ablation study and comparison with the state-of-the-art approaches. Our contributions can be summarized as follows:
\begin{itemize}
\item We propose a novel Diminishing Empirical Risk Minimization (DERM) framework for unsupervised anomaly detection. In DERM, the adverse effect of the potential anomalies are suppressed in a dynamic and controllable manner. 

\item We conduct theoretical analysis on DERM and reveal that DERM can directly modify the gradient contribution of each individual loss. Specifically, each gradient contribution is proportional to the ratio of the average loss to a single loss. 

\item We perform extensive experiments on both synthetic and real-world datasets to verify the efficacy of DERM. Experimental results demonstrate improved performance of DERM on the unsupervised AD benchmark consisting of 18 datasets.
\end{itemize}

\section{Related works}
\subsection{\textbf{Unsupervised Anomaly Detection (AD).}}
Anomaly detection is a significant study field in machine learning and data mining \cite{ref:ADsurvey,ref:AD_survey2,ref:deep-AD-review,ref:Deep-AD_survey}. Unsupervised anomaly detection does not require any data with labelling. The only assumption is that the number of normal data points is larger than the number of anomalies. A number of methods have been proposed for unsupervised AD~\cite{ref:ADsurvey,ref:Deep-AD_survey,ref:deep-AD-review}. 
Traditional methods tend to choose Principal Component Analysis (PCA) \cite{ref:PCA}, Isolation Forest \cite{ref:IF} and Support Vector Machine (SVM) \cite{ref:OC-SVM} to detect anomalies in an unsupervised manner.

Recently, a large amount of representation learning approaches equipped with deep neural network have aroused great interest in this space. The core idea is to learn an useful representation by minimizing the reconstruction loss. These 
reconstruction-based methods learn a low-dimensional vector in the latent space and then project it back to the original feature space. The reconstruction error between the input and the reconstructed output is treated as the anomaly score. 
Early work \cite{ref:outlier-ae} in this branch is proposed on anomaly detection with the representation learning capability of autoencoder, which utilized the large reconstruction error to detect anomalies. 
Subsequent work combined diverse techniques or prior knowledge with autoencoder to enhance the detection efficacy.
RDA \cite{ref:pca-ae} combined the  robust PCA with an autoencoder to group the data into a mixture of normal and anomaly components. 
DAGMM~\cite{ref:dagmm} trained a Gaussian Mixture Model to learn the latent representation from autoencoder to determine anomalies jointly by reconstruction error and density estimation. 
In addition, one-class classification and its deep neural network variants are also widely used for anomaly detection~\cite{ref:OC-SVM,ref:Deep-SVDD}. 
The decision boundary surrounding normal instances is also learned for anomaly detection. 

However, these methods often rely on the DNN to extract information, and most existing deep AD approaches fail to prevent the DNN from aggressively learning the anomalies. 
Recent studies attempted to address this issue via different strategies. 
RCA~\cite{ref:RCA} proposed to discard a proportion of suspicious anomaly data through a collaborative autoencoder. 
RSR~\cite{ref:rsr-ae} utilized a robust subspace recovery layer to extract a subspace from the given data and move the outliers further away from the subspace. 
Different from such algorithms, we resolve this issue from the learning principle perspective and propose the Diminishing Empirical Risk Minimization (DERM) framework. DERM adaptively adjusts the individual loss contribution of each instance to diminish the outliers. 


\subsection{\textbf{Autoencoders for Anomaly Detection}}
\textcolor{black}{Autoencoders have been widely-adopted in unsupervised anomaly detection~\cite{ref:dagmm,ref:un-ad-vae,ref:MemAE,ref:hpcAE,ref:industAE}. The general idea is to train an autoencoder on the entire dataset (both normal and anomalous) and utilize the reconstruction error as the detection criterion. As there are fewer anomalous data for training, the reconstruction errors of anomalies are higher than that of normal ones, which can be utilized to separate the anomalies from normal data. There are subsequent work that seek help from traditional machine learning algorithms to enhance the capability of vanilla autoencoder. For example, robust autoencoder \cite{ref:deep-inductive-AD} incorporates RPCA \cite{ref:robust-pca} in an autoencoder, where parameters of autoencoder and a sparse residual matrix are alternatively optimized. 
Normalized deep autoencoder~\cite{ref:normalized-deep-ae} considers the situation of multiple modes for normal instances and also applies $L_2$ normalization for latent variables of the autoencoder. What's more,
MemAE~\cite{ref:MemAE} proposes a memory-augmented autoencoder to improve the performance of unsupervised anomaly detection. Nevertheless,
most existing approaches overlook the deficiency of the default ERM principle in AD applications. In this work, we put forward a novel DERM framework to circumvent the drawback of ERM.
}
\subsection{\textbf{Sample Re-weighting and Aggregation Schemes.}}
Approaches have been proposed to re-weight the influence of samples by modifying the ERM objective. 
In~\cite{ref:reweight2,ref:reweight3}, examples were re-weighted as per their loss values to intervene the optimization dynamics, which pays more attention to difficult examples. 
\textit{Relaxed clipping} \cite{ref:loss-clipping} performed the example re-weighting via loss clipping. 
There are other works trying to modify the loss aggregation scheme. One of the alternatives to traditional average loss in ERM is to consider a min-max objective, which tries to minimize the max loss. The min-max objective has been applied in application such as enhancing robustness under perturbations~\cite{ref:min-max-application}. 

Sample re-weighting has also been applied on anomaly detection. Most of existing deep anomaly detection approaches fail to protect the neural network from interfering by anomalies during parameter learning. Some work address this issue via assigning different weights to corresponding data point. For instance, self-paced learning model \cite{ref:self-paced} and Mentornet \cite{ref:Mentornet} assign higher weights to instances which are easier to be classified. Recently, TERM~\cite{ref:TERM} proposed the tilted empirical risk minimization, which tuned the impact of individual losses by applying different gradient weights on them. 
However, existing methods are less effective when applied to unsupervised AD. 
By contrast, DERM is specially designed for the unsupervised AD task by adaptively re-weighting examples with a new learning principle. 
Moreover, DERM can directly modify the gradient contribution of each example from a theoretical perspective.




\begin{figure*}[t]
	\centering  
	\includegraphics[width=0.75\linewidth]{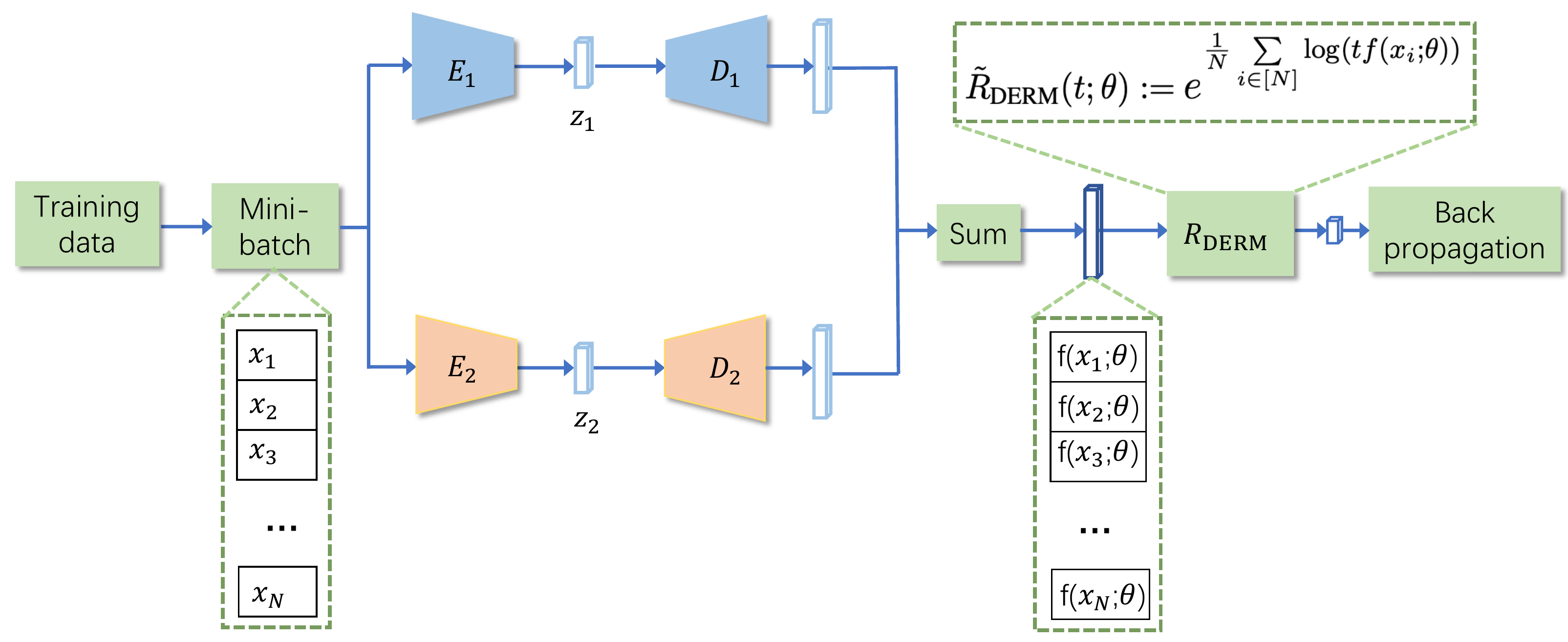}  
	\caption{\label{framework}Illustration of the DERM framework in the training phase. A batch of training data is sampled as the input for $k$ ($k=2$ in this figure) collaborative autoencoders. 
	The reconstruction loss for each instance is computed. 
	Then, the total loss for an instance from the two  autoencoders is summed to obtain a batch of loss. DERM aggregation scheme is applied on the loss and back propagation is performed to update  parameters in neural networks of both  autoencoders.} 
\end{figure*}

\section{Methodology}
\subsection{Diminishing Empirical Risk Minimization (DERM)}

\paragraph{\textbf{Problem Setting.}} 
Suppose we are given a dataset $X = \{x_1,x_2,\dots,x_n\}$, where $x_i \in X \subseteq \mathbb{R}^{d} $ has $d$ features. Our objective is to classify each $x_i$ as either an anomalous instance or a normal one in an unsupervised manner.



\paragraph{\textbf{DERM Framework.}}
DERM learns a set of $k$ autoencoders $\{(E_j,D_j)|j=1,2 \dots k\}$ with different initializations. Here, we set $k=2$ for brevity (the discussion be easily generalized to the situation when $k>2$). 
As shown in Figure~\ref{framework}, in each iteration, a mini-batch of data is randomly sampled and fed into the autoencoders. Then, the reconstruction loss for each data instance $x_i$ is calculated as: 
\begin{equation}
    f(x_i;\theta) = \sum\limits_{j = 1,2,\dots,k} ||x_i - D_j(E_j(x_i))||_2^2\,.
\end{equation}

To obtain the final loss, as discussed before, previous work utilize the  ERM~(Eq.~\ref{eq:erm}) to aggregate all the instance losses with equal weights. 
However, it ignores the distinctions between the normal and anomalous instances, thus overfitting the anomalies. 
To address this issue, we devise the DERM (Eq.~\ref{eq:DERM}) to automatically and dynamically adjust the loss weights for each instance. With the DERM, the normal instances are well learned and result in small losses, while the anomalous instances are less-trained and cause large losses due to their inconsistent behaviour and the lack of data. 
Then, a back-propagation step is conducted to update the parameters of the whole model by gradient-based optimization. 
During the testing phase, the reconstruction loss $f(x_i;\theta)$ is adopted as an anomaly score of the testing instance. The details of the DERM framework (k=2) is shown in Algorithm~\ref{algo:DERM}.

\paragraph{\textbf{Theoretical Analysis.}} To dig deep into the motivation and mechanism, we present the theoretical analysis for the proposed DERM framework. 
Using traditional ERM is equivalent to obtaining the mean loss of the training samples, which can be biased towards (or negatively affected by) outlier data. 
By the following Theorem, we show that DERM can suppress the anomalies by reducing the gradient contribution of the potential anomalous candidates. 

\newtheorem{theorem}{Theorem}
\begin{theorem}
\label{theorem:1}
 For a continuously differentiable (i.e., smooth) loss function $f(x;\theta)$, the gradient of DERM (Eq.~\ref{eq:DERM}) is: 
\begin{equation}
\nonumber
\nabla_{\theta} \tilde{R}_{\text{DERM}}(t;\theta) =
\sum\limits_{i \in [N]}^{} w_i(t;\theta) \nabla_{\theta}  f(x_i;\theta) \, \end{equation}
\begin{equation}
\nonumber
\text{ where }\,
w_i(t;\theta)
= \frac{1}{N}  \frac{ \tilde{R}_{\text{DERM}}(t;\theta)}{f(x_i;\theta)}.
\nonumber
\end{equation}
\end{theorem}

\newtheorem{proof}{Proof}
\begin{proof}
\begin{align}
\nonumber
& \nabla_{\theta} \tilde{R}_{\text{DERM}}(t;\theta) \\
\nonumber
& = \nabla_{\theta} \left\{ 
e^{\frac{1}{N} \sum\limits_{i \in [N]} \log  t f(x_i;\theta)} 
\right\} \\
\nonumber
& = \left(  e^{\frac{1}{N} \sum\limits_{i \in [N]} \log  t f(x_i;\theta)}  \right)
\boldsymbol{\cdot} \frac{1}{N} \sum\limits_{i \in [N]}\nabla_{\theta}  \log  tf(x_i;\theta) \\
\nonumber
& = \left(e^{\frac{1}{N} \sum\limits_{i \in [N]} \log  tf(x_i;\theta) } \right)
\boldsymbol{\cdot} \frac{1}{N} \sum\limits_{i \in [N]} \frac{\nabla_{\theta}  f(x_i;\theta)}{f(x_i;\theta)} \\
\nonumber
& = \sum\limits_{i \in [N]}^{}  \frac{1}{N} \boldsymbol{\cdot} \frac{
e^{\frac{1}{N} \sum\limits_{i \in [N]} \log  tf(x_i;\theta) 
} }{f(x_i;\theta)} \nabla_{\theta}  f(x_i;\theta) \\
\nonumber
& = \sum\limits_{i \in [N]}  \frac{1}{N} \boldsymbol{\cdot} \frac{ \tilde{R}_{\text{DERM}}(t;\theta)}{f(x_i;\theta)} \nabla_{\theta}  f(x_i;\theta) 
\end{align} 
\IEEEQED
\end{proof}
%


We can observe from Theorem~\ref{theorem:1} that the gradient of DERM is a weighted sum of the individual gradient terms w.r.t the instances. 
Specifically, for each instance $x_i$, the original 
gradient $\nabla_{\theta}  f(x_i;\theta)$ is re-weighted proportional to the ratio between the DERM loss $\tilde{R}_{\text{DERM}}(t;\theta)$ and the instance loss $f(x_i;\theta)$. 
Considering the gradient ratio structure in Theorem~\ref{theorem:1}, the DERM term in the numerator is fixed in a mini-batch, while the denominator $f(x_i; \theta)$ is a strong indicator for the anomalous/normal probability. 
In particular, a normal instance usually has a small loss $f(x_i; \theta)$, leading to a large re-weighting ratio on the original gradient. On the contrary, an anomalous instance tends to have a large loss, thus resulting in a small ratio on the original gradient. 
With this mechanism, the influence of the anomalous instances is weakened while the impact of the normal instances are reinforced, which alleviates the overfitting and universal approximation towards  the anomalies. 

Compared with the existing methods that only change the weights of the instance losses, our method directly re-weights the gradient terms to intervene the optimization process, which works on the parameter space in a more straightforward manner. Moreover, taking both the batch statistics (i.e., $\tilde{R}_{\text{DERM}}(t;\theta)$) and the individual loss $f(x_i; \theta)$ into consideration, DERM can automatically and dynamically modify the gradient weights. This avoids designating manual or ad-hoc weights according to the dataset, thus facilitating the generalization of our method to a diverse range of datasets without many extra adjustments. 

\begin{algorithm}[t]
 \KwIn{training set $X_{train}$, test set $X_{test}$, temperature $t$, 
 learning rate $\eta$, max epoch $M$.}
  \KwOut{anomaly scores for $X_{test}$.}
  \textbf{Initialize} encoders $E_1$, $E_2$ and decoders $D_1$, $D_2$ with $\theta = \{\theta_{E_1}, \theta_{E_2}, \theta_{D_1}, \theta_{D_2} \}$.\\
  
  \textbf{Training phase:} \\
  
  \For{epoch $\in \{1,2,\dots M\}$}{
  Sample a mini-batch $\{x_1,\dots,x_N\}$ from $X_{train}.$ \\
  Calculate $f(x_i;\theta) = ||x_i - D_1(E_1(x_i)) ||_2^2 + ||x_i - D_2(E_2(x_i)) ||_2^2$ for each $x_i$ in the sampled mini-batch.\\
  
  Aggregate all losses $\{ f(x_i;\theta) \}_N$ with $\tilde{R}_{\text{DERM}}(t;\theta)$ according to Eq.~\ref{eq:DERM}.\\
  
  Update $\theta \leftarrow \theta - \eta \nabla_{\theta} \tilde{R}_{\text{DERM}}(t;\theta).$
  }
  
  \textbf{Testing phase:} \\
  \For{each $x_i$ in $X_{test}$ }{
     Calculate anomaly score $s(i) = f(x_i;\theta)$ for $x_i$.
    }
  
  return anomaly scores $s$.
  \caption{\label{algo:DERM}DERM for unsupervised AD}
\end{algorithm}


\subsection{Comparison with Existing Methods} 

Studies have been carried out to devise alternatives to the vanilla ERM in various scenarios, such as classification with imbalanced data~\cite{ref:focal_loss,ref:imbalance} and  learning in the presence of corrupted data~\cite{ref:corrupted_label,ref:noise}.  
Recently, tilted empirical risk minimization (TERM)~\cite{ref:TERM} framework is proposed to address the sensitivity and poor generalization issues in ERM. 
TERM takes the following form:
\begin{equation}
\label{eq:term0}
\tilde{R}_{\text{TERM}}(t;\theta) := \frac{1}{t}
\log \left( \frac{1}{N} \sum\limits_{i \in [N]} e^{ tf(x_i;\theta)} \right).
\end{equation}

TERM is shown to be effective in robust regression and classification tasks. However, it turns out to be less effective on unsupervised anomaly detection due to its intrinsic property. 
TERM suffers from the sensitivity to the absolute value of loss, resulting in unsatisfactory outcomes. We  analyze the problem as follows, as a comparison to our DERM in 
Theorem~\ref{theorem:1}. The gradient of TERM \cite{ref:TERM} is,
\begin{equation}
\label{eq:term}
\fontsteqn \nabla_{\theta} \tilde{R}_{\text{TERM}}(t;\theta) =
\sum\limits_{i \in [N]}^{} w_i(t;\theta) \nabla_{\theta}  f(x_i;\theta) \end{equation}
\begin{equation}
\text{ with }
w_i(t;\theta)
= \frac{e^{t(f(x_i;\theta)-\tilde{R}_{\text{TERM}}(t;\theta))}}{N}\,.
\end{equation}

When applying TERM to the unsupervised anomaly detection, we find a major drawback with respect to Eq.~\ref{eq:term} (note that when applying TERM to anomaly detection, it requires $t<0$ \cite{ref:TERM}.). 
For a min-batch of data $X = \{x_1,x_2,...,x_N\}$, according to DNN's universal approximation property, the reconstruction loss $\{f(x_i;\theta)\}$ has a high possibility to distribute close to $0$. 
Consequently, the item $|f(x_i;\theta) - \tilde{R}_{\text{TERM}}|$ for a data point $x_i$ is also highly likely to approach $0$ since $\tilde{R}_{\text{TERM}}$ is a variant of average w.r.t $X$. 
This leads to an unsatisfactory trivial solution that the gradient weight $w_i(t;\theta)$ of each instance in Eq.~\ref{eq:term} is close to $\frac{1}{N}$. 
To amend the issue, TERM heavily relies on the parameter $t$ to scale $|f(x_i;\theta) - \tilde{R}_{\text{TERM}}|$ into a suitable range. Unfortunately, $X$ often changes drastically for different instances even in the same dataset. 
In realistic applications, it takes efforts to determine the ideal value of $t$ and sometimes the ideal $t$ does not exists. 
If we want to apply the same algorithm to multiple datasets, the sensitivity of $t$ requires frequently re-training the model and hinders the knowledge transferring across datasets, which restricts its application scenario. 

The proposed DERM avoids the aforementioned issue in TERM. 
Specifically, in DERM, $w_i(t;\theta)$ is linear to $\frac{ \tilde{R}_{\text{DERM}}(t;\theta)}{f(x_i;\theta)}$. 
By contrast, in TERM, $w_i(t;\theta)$ corresponds to $e$ to the power of $(f(x_i;\theta) - \tilde{R}_{\text{TERM}})$. 
This formulation difference leads to following advantage of our proposed method: 
the weight $w_i(t;\theta)$ of an instance $x_i$ is not sensitive to the distribution and numerical value of training data since it takes the ratio format instead of the subtraction between $\tilde{R}_{\text{DERM}}$ and $f(x_i;\theta)$. This avoids trivial solution discussed above and assigns discriminative weights to normal and anomalous instances, respectively.
Moreover, the form of $w_i(t;\theta)$ has an additional signal magnification effect. 
Except for the anomalous instances, the normal instances can be dynamically weighted to reflect their various impacts. 
This way, DERM makes use of the pattern of normal data to emphasize more on high-quality normal data. 


\begin{figure}[t]
	\centering  
	\includegraphics[width= 1\linewidth]{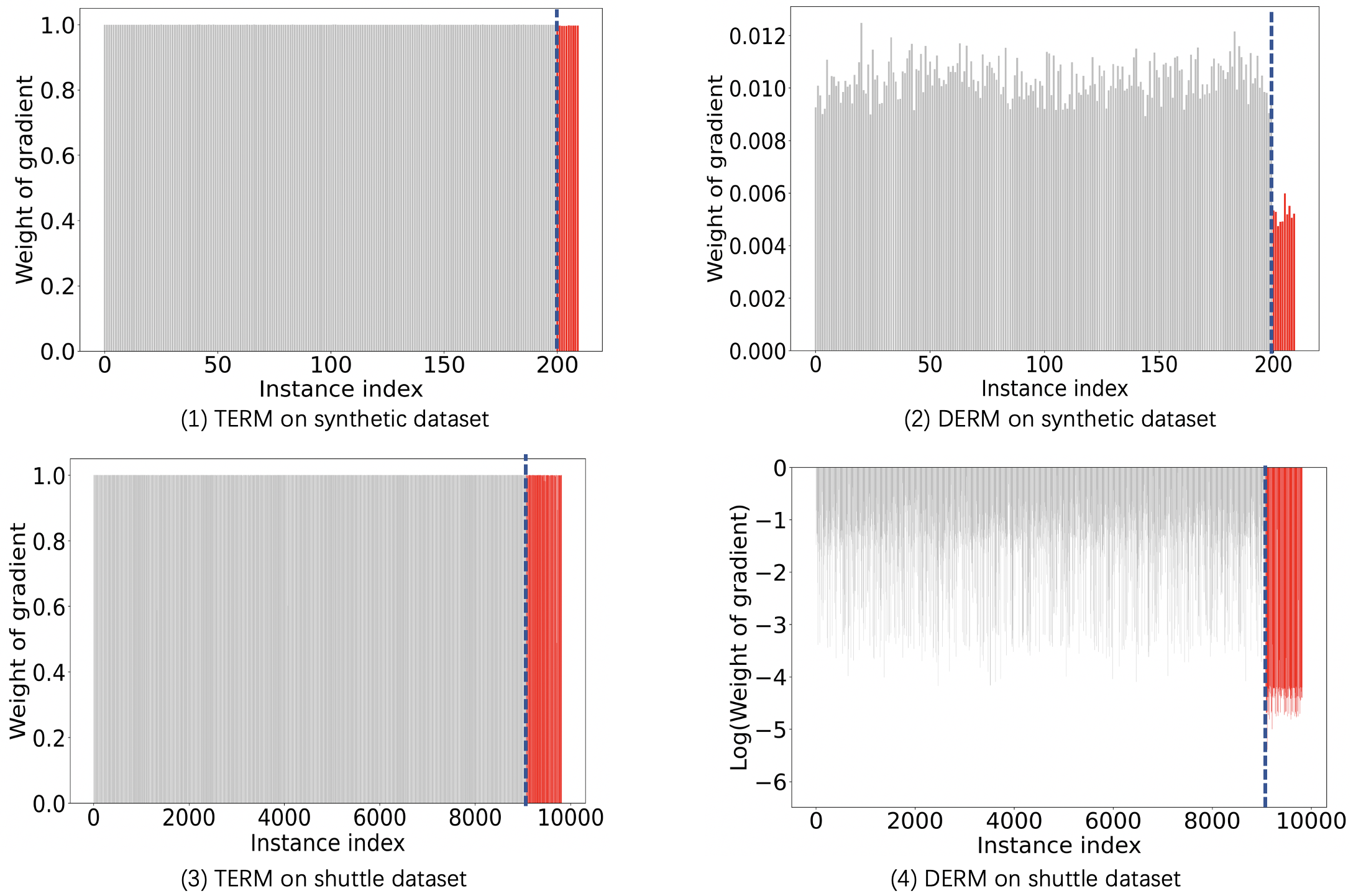}
	\caption{Gradient comparison between DERM and TERM~\cite{ref:TERM} on real and synthetic datasets. The height of grey bar represents the gradient weights of the normal instances. The height of red bar represents that of anomalous instances. The gradient weight is obtained by $\omega_i/{\frac{1}{N}}$ for clear visualization.
	Blue dash line separates the normal and anomalous data. 
	In (4), a base-10 logarithm scale is applied on the y-axis for a better visualization. 
	}
	\label{fig:compare_gradients}
\end{figure}


To better understand how DERM overcomes the limitations of TERM, we conduct experiments on both synthetic and real-world dataset. For synthetic dataset, we generate reconstruction loss for 200 normal instances from $\mathcal{N}(0.03,0.002)$ and 10 anomalies from $\mathcal{N}(0.06,0.005)$ that constitute a set $\{ f(x_i;\theta)\}_{i\in [210]}$ to simulate the losses for both normal and anomalous data. For real-world dataset, we adopt the $shuttle$ dataset from OODS library~\cite{ref:oods}. 
The gradient weights of all instances are shown in Figure ~\ref{fig:compare_gradients},  
%
from which we can draw several remarks as follows: (1) the gradient weights of TERM approach $\frac{1}{N}$, since the exponential powers in Eq.~\ref{eq:term} are close to $0$, 
(2) normal and anomalous instances are better distinguished and separated by the proposed DERM method, (3) except for the normal/anomalous instance separation, DERM also induces dynamic weights on the normal instances. These remarks are consistent with the above analysis and demonstrates our DERM method as a better anomaly detector. 



\begin{table*}[t]
\centering
\caption{\label{tab:performace}AUC values (mean$\pm$std) on 18 datasets across diverse domains.}
\setlength{\tabcolsep}{2.0pt}
\resizebox{0.8\textwidth}{!}{
\begin{tabular}{l|llllllllll} 
\hline  
& \fontstdd RCA & \fontstdd VAE & \fontstdd SO\_GAAL & \fontstdd RSR & \fontstdd DAGMM & \fontstdd D-SVDD & \fontstdd OCSVM & \fontstdd IF & \fontstdd DERM(Ours)  \\
\cline{1-10}
vowels      & 0.917\fontstd$\pm$.016          & 0.503\fontstd$\pm$.045          & 0.555\fontstd$\pm$.219          & 0.930\fontstd$\pm$.019          & 0.340\fontstd$\pm$.103 & 0.190\fontstd$\pm$.062 & 0.767\fontstd$\pm$.044          & 0.765\fontstd$\pm$.031          & \textbf{0.972}\fontstd$\pm$.011 &   \\
pima        & \textbf{0.711}\fontstd$\pm$.016 & 0.648\fontstd$\pm$.015          & 0.629\fontstd$\pm$.054          & 0.660\fontstd$\pm$.050          & 0.531\fontstd$\pm$.025 & 0.363\fontstd$\pm$.051 & 0.633\fontstd$\pm$.016          & 0.673\fontstd$\pm$.012          & 0.704\fontstd$\pm$.017          &   \\
optdigits   & 0.890\fontstd$\pm$.041          & 0.909\fontstd$\pm$.016          & 0.495\fontstd$\pm$.185          & 0.885\fontstd$\pm$.180          & 0.290\fontstd$\pm$.042 & 0.550\fontstd$\pm$.059 & 0.555\fontstd$\pm$.039          & 0.726\fontstd$\pm$.049          & \textbf{0.893}\fontstd$\pm$.036 &   \\
sensor      & 0.950\fontstd$\pm$.030          & 0.913\fontstd$\pm$.003          & 0.698\fontstd$\pm$.238          & 0.980\fontstd$\pm$.013          & 0.924\fontstd$\pm$.085 & 0.614\fontstd$\pm$.100 & 0.941\fontstd$\pm$.002          & 0.949\fontstd$\pm$.009          & \textbf{0.996}\fontstd$\pm$.001 &   \\
letter      & 0.802\fontstd$\pm$.036          & 0.521\fontstd$\pm$.042          & 0.414\fontstd$\pm$.094          & 0.665\fontstd$\pm$.045          & 0.433\fontstd$\pm$.034 & 0.465\fontstd$\pm$.073 & 0.531\fontstd$\pm$.060           & 0.638\fontstd$\pm$.021          & \textbf{0.829}\fontstd$\pm$.022 &   \\
cardio      & 0.905\fontstd$\pm$.012          & \textbf{0.944}\fontstd$\pm$.006  & 0.449\fontstd$\pm$.105          & 0.937\fontstd$\pm$.009          & 0.862\fontstd$\pm$.031 & 0.505\fontstd$\pm$.047 & 0.932\fontstd$\pm$.008          & 0.930\fontstd$\pm$.008          & 0.886\fontstd$\pm$.021          &   \\
arrhythmia  & 0.806\fontstd$\pm$.044          & 0.811\fontstd$\pm$.034          & 0.558\fontstd$\pm$.077          & 0.893\fontstd$\pm$.006          & 0.603\fontstd$\pm$.095 & 0.635\fontstd$\pm$.065 & 0.811\fontstd$\pm$.061          & 0.807\fontstd$\pm$.007          & \textbf{0.898}\fontstd$\pm$.002 &   \\
breastw     & 0.978\fontstd$\pm$.003          & 0.950\fontstd$\pm$.006          & 0.985\fontstd$\pm$.011          & 0.956\fontstd$\pm$.007          & 0.976\fontstd$\pm$.000 & 0.406\fontstd$\pm$.047 & 0.955\fontstd$\pm$.015          & \textbf{0.987}\fontstd$\pm$.002 & 0.951\fontstd$\pm$.003          &   \\
musk        & \textbf{1.000}\fontstd$\pm$.000 & 0.944\fontstd$\pm$.002          & 0.840\fontstd$\pm$.218          & 0.964\fontstd$\pm$.002          & 0.903\fontstd$\pm$.130 & 0.829\fontstd$\pm$.072 & \textbf{1.000}\fontstd$\pm$.000 & 0.998\fontstd$\pm$.004          & \textbf{1.000}\fontstd$\pm$.000 &   \\
mnist       & \textbf{0.858}\fontstd$\pm$.012 & 0.778\fontstd$\pm$.009          & 0.767\fontstd$\pm$.058          & 0.754\fontstd$\pm$.065          & 0.652\fontstd$\pm$.077 & 0.538\fontstd$\pm$.069 & 0.820\fontstd$\pm$.012          & 0.796\fontstd$\pm$.014          & 0.803\fontstd$\pm$.018          &   \\
satimage-2  & 0.977\fontstd$\pm$.008          & 0.966\fontstd$\pm$.008          & 0.772\fontstd$\pm$.158          & \textbf{1.000}\fontstd$\pm$.000 & 0.853\fontstd$\pm$.113 & 0.739\fontstd$\pm$.137 & 0.999\fontstd$\pm$.002          & 0.993\fontstd$\pm$.001          & 0.997\fontstd$\pm$.001          &   \\
satellite   & \textbf{0.712}\fontstd$\pm$.011 & 0.538\fontstd$\pm$.016          & 0.634\fontstd$\pm$.049          & 0.649\fontstd$\pm$.048          & 0.667\fontstd$\pm$.189 & 0.631\fontstd$\pm$.023 & 0.653\fontstd$\pm$.014          & 0.702\fontstd$\pm$.021          & 0.661\fontstd$\pm$.041          &   \\
mammo & 0.844\fontstd$\pm$.014          & \textbf{0.864}\fontstd$\pm$.014 & 0.232\fontstd$\pm$.005          & 0.769\fontstd$\pm$.028          & 0.834\fontstd$\pm$.000 & 0.272\fontstd$\pm$.048 & 0.830\fontstd$\pm$.027          & 0.862\fontstd$\pm$.010          & 0.801\fontstd$\pm$.041 & \\
thyroid     & 0.956\fontstd$\pm$.008          & 0.839\fontstd$\pm$.011          & \textbf{0.984}\fontstd$\pm$.032 & 0.940\fontstd$\pm$.023          & 0.582\fontstd$\pm$.095 & 0.704\fontstd$\pm$.076 & 0.893\fontstd$\pm$.026          & 0.979\fontstd$\pm$.003          & 0.951\fontstd$\pm$.020          &   \\
annthyroid  & 0.688\fontstd$\pm$.016          & 0.589\fontstd$\pm$.021          & 0.640\fontstd$\pm$.033          & 0.646\fontstd$\pm$.024          & 0.506\fontstd$\pm$.020 & 0.591\fontstd$\pm$.030 & 0.597\fontstd$\pm$.023          & \textbf{0.827}\fontstd$\pm$.011 & 0.692\fontstd$\pm$.027          &   \\
ionosphere         & 0.846\fontstd$\pm$.015          & 0.763\fontstd$\pm$.015          & 0.838\fontstd$\pm$.043          & 0.946\fontstd$\pm$.019          & 0.467\fontstd$\pm$.082 & 0.735\fontstd$\pm$.074 & 0.838\fontstd$\pm$.056          & 0.853\fontstd$\pm$.006          & \textbf{0.977}\fontstd$\pm$.016 &   \\
pendigits   & 0.856\fontstd$\pm$.011          & 0.931\fontstd$\pm$.006          & 0.272\fontstd$\pm$.062          & 0.884\fontstd$\pm$.057          & 0.872\fontstd$\pm$.068 & 0.613\fontstd$\pm$.052 & \textbf{0.957}\fontstd$\pm$.007 & 0.950\fontstd$\pm$.015          & 0.866\fontstd$\pm$.023          &   \\
shuttle     & 0.935\fontstd$\pm$.013          & 0.987\fontstd$\pm$.001          & 0.715\fontstd$\pm$.310          & 0.989\fontstd$\pm$.003          & 0.890\fontstd$\pm$.109 & 0.531\fontstd$\pm$.260 & 0.984\fontstd$\pm$.003          & \textbf{0.997}\fontstd$\pm$.001 & 0.981\fontstd$\pm$.001          &   \\ 
\cline{1-10}
Average     & 0.868                & 0.800                  & 0.638                & 0.859                & 0.676       & 0.539       & 0.808                & 0.855                & \textbf{0.881}       &   \\
\cline{1-10}
\end{tabular}}
\end{table*}

\subsection{Collaborative Autoencoders (cAE)}
Employing only one autoencoder, the model has the risk of quickly converging to an unsatisfactory solution due to the low reconstruction loss to compute the gradient~\cite{ref:general}. 
The premature convergence of loss function fails to explore the loss surface sufficiently~\cite{ref:RCA}. 
Meanwhile, ensemble of model outputs has shown a high efficacy in previous studies \cite{ref:IF,ref:meta-analysis,ref:outlier-ensemble,ref:pyod}.
To alleviate above issue, inspired by ensemble, we propose the Collaborative Autoencoders (cAE) with different weight initializations for each autoencoder.

Typically, ensemble is a collaboration of a set of models that are trained individually. 
To enable the end-to-end model training in DERM, we utilize a diagram of optimizing multiple autoencoders in parallel to mimic the traditional ensemble procedure. 
That is, all autoencoders are optimized by gradient based methods simultaneously and then collaboratively contribute to the subsequent testing phase. After training, each individual autoencoder would have distinct parameters as they are initialized with random values. Subsequently, an one forward pass is performed over the data to obtain multiple reconstruction losses for each test data point. The final loss is a summation of all reconstruction losses, which reduces the chance of model becoming overfitting to certain data points. This collaboration of multiple AEs design endows more robustness and reduce the testing variance, which further enhances the entire performance for detecting anomalies.
Collaborative or ensemble structure has also been adopted in a recent work~\cite{ref:RCA}, which performs two forward passes of each autoencoder, resulting in an increased computation cost. It also adopts dropouts to simulate ensemble process, which can bring unstability and undermine the performance.
In contrast, our proposed cAE structure requires only one forward pass of each autoencoder. There is no dropout in proposed structure, thus avoiding potential negative effect on model capability.


\section{Experimental Results}
To evaluate the proposed DERM framework, we perform experiments on real-world outlier detection datasets from diverse domains with both continuous and categorical features. 
We follow the setting in \cite{ref:RCA} to carry out experiments on 18 datasets from the OODS library \cite{ref:oods}, on which previous algorithms also perform experiments. Specifically, we split data such that 80\% is used for training, and the remaining 20\% for testing. 

For a single autoencoder, both the encoder and decoder are implemented by multi-layer feedforward neural networks with two hidden layers. The model is optimized with Adam optimizer with a default learning rate of 0.001. The setting for training is consistent with all DNN-based benchmarks. 
Our method is not sensitive to the parameter $t$ (as will be discussed in Section 4.2 and shown in Fig.~\ref{fig:sensitivity}), we thus empirically set it to 0.01 for all datasets. 
The commonly-used Area under ROC curve (AUC) score is adopted as the evaluation metric for all methods.

\subsection{Benchmarks and settings}


We compare DERM with the following benchmarks:
\begin{itemize}
\item RCA~\cite{ref:RCA}, which is a recent state-of-the-art robust framework using collaborative autoencoders to jointly identify normal observations from the data while learning its feature representation.
\item Variational AutoEncoder (VAE)~\cite{ref:vae}, which is a probabilistic model aiming to learn a Bayesian latent variable model by maximizing the log-likelihood of the training data.
\item SO-GAAL~\cite{ref:SO-GAAL}, which is a novel Single-Objective Generative Adversarial Active Learning method. It directly generates informative potential outliers based on the mini-max game between a generator and a discriminator.
\item RSR~\cite{ref:rsr-ae}, which is a neural network model with a novel robust subspace recovery layer. This layer extracts the underlying subspace from a latent representation of the given data and removes outliers that lie away from this subspace.
\item DAGMM~\cite{ref:dagmm}, which trains a Gaussian Mixture Model to learn the latent representation from the autoencoder to determine anomalies jointly by the reconstruction error and the density estimation.
\item Deep-SVDD~\cite{ref:Deep-SVDD}, which learns a neural network transformation from input space to output space. The transformation attempts to map most of the data representations into a hyper-sphere with radius of minimum volume.
\item OCSVM~\cite{ref:OC-SVM}, which is based on the  one-class SVM and fits a tight hyper-sphere in the non-linearly transformed feature space to include most of the data based on the positive examples.
\item Isolation Forest (IF)~\cite{ref:IF}, which builds an ensemble of trees for a given dataset. Anomalies are then identified as instances that have short average path lengths on the trees.
\end{itemize}

Among them, OCSVM and IF are traditional AD detectors, others are recently proposed  DNN-based approaches. 
For all the compared methods, we adopt the optimal settings from their official implementations with minimal modifications so that they can adapt to the datasets from OODS library. To ensure the fair comparison, the same neural network architecture is applied for all DNN-based algorithms. Experiments are repeated for 20 times with random initializations and the average$\pm$std are reported.

\subsection{Comparison with the State-of-the-Art}

\begin{table}[t]
\caption{\label{tab:ablation}Ablation study of our proposed method (DERM + cAE). cAE represents collaborative AEs with MSE loss and MSE anomaly score. In all cAE, number of AEs $k$ is set to 2. }
\centering
\setlength{\tabcolsep}{3.5pt}
\resizebox{0.45\textwidth}{!}{
\begin{tabular}{l|llll} 
\hline
            & \fontstddd Autoencoder             & \fontstddd TERM+cAE             & \fontstddd cAE         & \fontstddd DERM+cAE          \\ 
\hline
vowels      & 0.919\fontstd$\pm$.034 & 0.950\fontstd$\pm$.022          & 0.949\fontstd$\pm$.025 & \textbf{0.972}\fontstd$\pm$.011           \\
pima        & \textbf{0.704}\fontstd$\pm$.017 & 0.659\fontstd$\pm$.030          & 0.672\fontstd$\pm$.029 & 0.625\fontstd$\pm$.048           \\
optdigits   &0.854\fontstd$\pm$.065        & \textbf{0.927}\fontstd$\pm$.023 & 0.889\fontstd$\pm$.028 &0.893\fontstd$\pm$.036           \\
sensor      &0.960\fontstd$\pm$.015  & 0.975\fontstd$\pm$.026          & 0.970\fontstd$\pm$.020 &\textbf{0.996}\fontstd$\pm$.001            \\
letter      & \textbf{0.829}\fontstd$\pm$.022 & 0.822\fontstd$\pm$.016          & 0.796\fontstd$\pm$.040 & 0.819\fontstd$\pm$.054           \\
cardio      & \textbf{0.886}\fontstd$\pm$.021 & 0.867\fontstd$\pm$.025          & 0.868\fontstd$\pm$.031 & 0.878\fontstd$\pm$.036           \\
arrhythmia  &0.886\fontstd$\pm$.003  & 0.890\fontstd$\pm$.010          & 0.885\fontstd$\pm$.002 &\textbf{0.898}\fontstd$\pm$.002            \\
breastw     &0.933\fontstd$\pm$.016  &  0.942\fontstd$\pm$.007          & 0.939\fontstd$\pm$.006 &\textbf{0.951}\fontstd$\pm$.003           \\
musk        & 0.977\fontstd$\pm$.085 & \textbf{1.000}\fontstd$\pm$.001 & 0.989\fontstd$\pm$.022 &\textbf{1.000}\fontstd$\pm$.000           \\
mnist       & 0.780\fontstd$\pm$.016          & \textbf{0.850}\fontstd$\pm$.017 & 0.781\fontstd$\pm$.027 &
0.803\fontstd$\pm$.018\\
satimage-2  &0.923\fontstd$\pm$.027  & 0.942\fontstd$\pm$.027          & 0.960\fontstd$\pm$.022 &
\textbf{0.997}\fontstd$\pm$.001\\
satellite   &0.657\fontstd$\pm$.021           & \textbf{0.664}\fontstd$\pm$.015 & 0.654\fontstd$\pm$.015 &
0.661\fontstd$\pm$.041\\
mammo & \textbf{0.833}\fontstd$\pm$.033          & 0.828\fontstd$\pm$.030          & 0.832\fontstd$\pm$.028 &0.801\fontstd$\pm$.041   \\
thyroid     & 0.907\fontstd$\pm$.038 & 0.921\fontstd$\pm$.031          & 0.936\fontstd$\pm$.025 & \textbf{0.951}\fontstd$\pm$.020         \\
annthyroid  &0.656\fontstd$\pm$.023  & 0.671\fontstd$\pm$.019          & 0.667\fontstd$\pm$.017 &
\textbf{0.692}\fontstd$\pm$.027\\
ionosphere         &0.969\fontstd$\pm$.008           & \textbf{0.984}\fontstd$\pm$.002 & 0.968\fontstd$\pm$.007 &0.977\fontstd$\pm$.016            \\
pendigits   &0.799\fontstd$\pm$.047  & 0.803\fontstd$\pm$.040          & 0.826\fontstd$\pm$.043 &\textbf{0.866}\fontstd$\pm$.023            \\
shuttle     &0.838\fontstd$\pm$.144  & 0.812\fontstd$\pm$.080          & 0.813\fontstd$\pm$.059 &\textbf{0.981}\fontstd$\pm$.001            \\ 
\hline
Average     &0.845        & 0.860                 & 0.855       & \textbf{0.881}                 \\
\hline
\end{tabular}}
\end{table}


We show the comparison results on 18 datasets in Table~\ref{tab:performace}. 
DERM achieves the best average AUC and has the most number of best-performing datasets (i.e., 7 datasets). 
This demonstrates the effectiveness of DERM framework for unsupervised anomaly detection. 
The DNN-based methods such as DAGMM, Deep-SVDD and SO-GAAL fail to perform well when dealing with  contaminated data. 
As analyzed, the reason is that most DNN-based methods aggressively fit the anomalies and learn inaccurate features from them, which are supposed to be learned from normal data. 
The ERM scheme widely-used in these methods exacerbates this issue. 
RCA is also inferior to the proposed method because RCA arbitrarily discards the suspicious instances, which impairs the detection capacity.

\subsection{Ablation Study and Parameter Analysis}

In order to validate the effectiveness of the DERM principle, we conduct ablation study on DERM in comparison with AE and TERM, as shown in Table~\ref{tab:ablation}.
All datasets are divided into training and testing data with a ratio of 0.8:0.2. The baseline is an autoencoder with mean square error loss and mean square error anomaly score, which is usually the default choice for autoencoder. The result clearly shows the advanced  performance of DERM over TERM and vanilla Autoencoder, validating the effectiveness of the DERM principle and the cAE design.

\begin{figure}[htb]
	\centering  
	\includegraphics[width=1\linewidth]{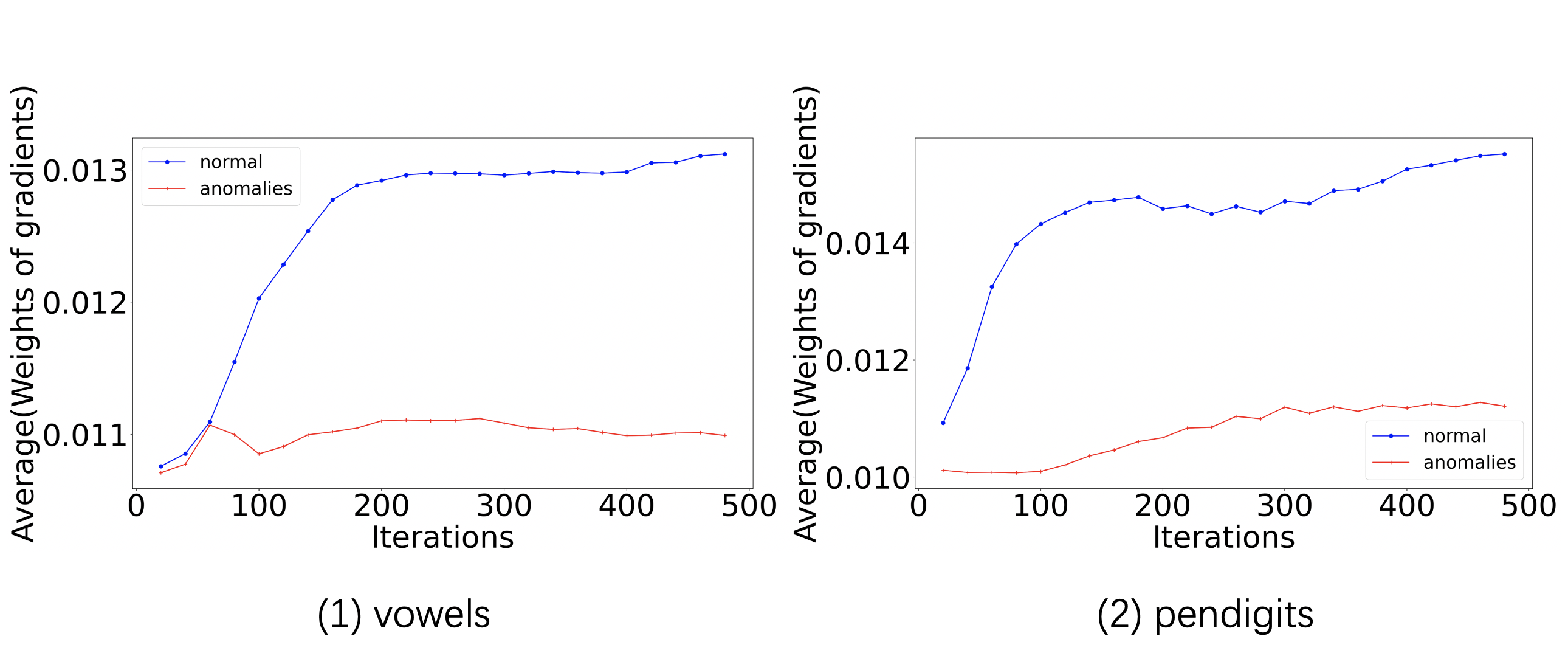}  
	\caption{Change of average weight of gradients for normal (blue) and anomalous (red) instances in $vowels$ and $pendigits$ dataset w.r.t. training iterations in DERM framework. } 
	\label{fig:weights_change}
\end{figure}

To better understand the effectiveness of the training dynamics for the proposed DERM principle, we compute and plot the change of average weight of gradients for all normal and anomalous instances in $vowels$ and $pendigits$ datasets w.r.t. training iterations, as shown in Fig.~\ref{fig:weights_change}. 
It clearly demonstrates that the weights of anomalies are almost consistently suppressed along the training process, which ensures the stability of optimization.

\begin{figure}[htb]
	\centering  
	\includegraphics[width=0.75\linewidth]{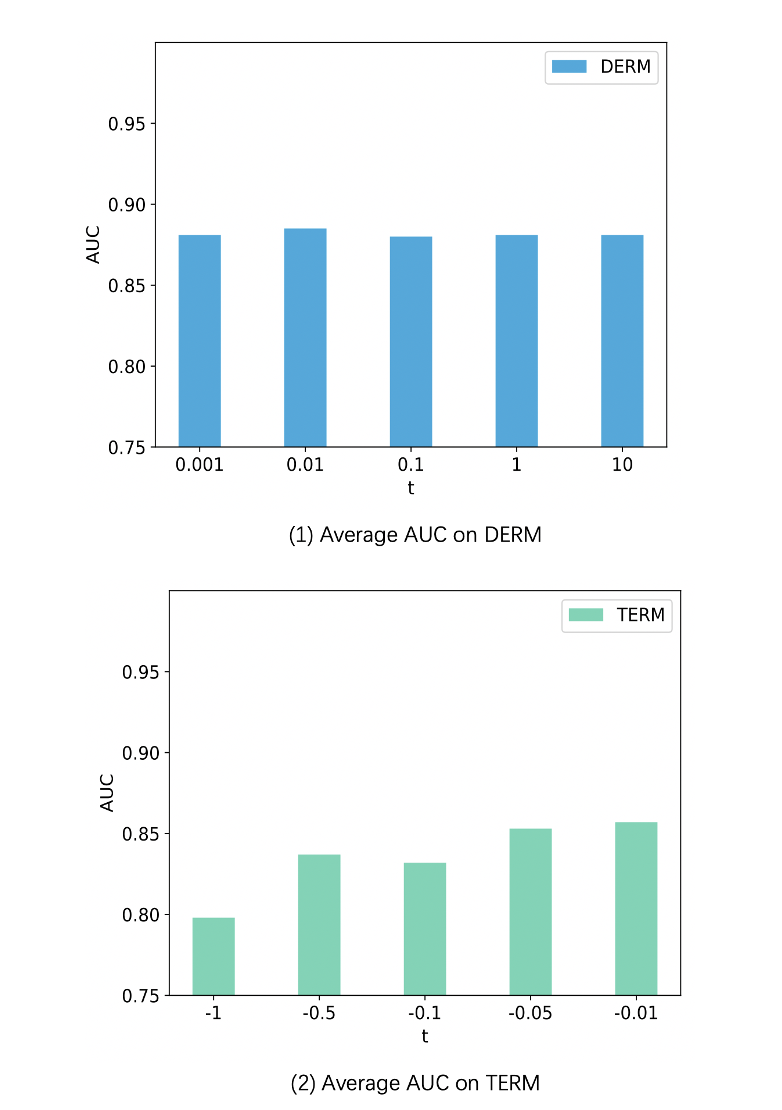}  
	\caption{Mean AUCs among 18 datasets with various $t$ via DERM and TERM respectively. Note that when  applying TERM  to anomaly  detection, it  requires $t < 0$. Hence, different ranges of $t$ are chosen for DERM and TERM for the sensitivity analysis. DERM shows robustness with even a larger variation of $t$.}
	\label{fig:sensitivity}
\end{figure}

Fig.~\ref{fig:sensitivity} shows the sensitivity analysis for mean AUCs among 18 datasets used in Fig.~\ref{tab:performace} by varying $t$ from 0.001 to 10 for DERM and -1 to -0.01 for TERM respectively. It can be observed that DERM is generally insensitive to the variation of parameters $t$ compared to TERM. It further validates the low testing variance and high robustness of DERM.


\section{Conclusion}

We propose a Diminishing Empirical Risk Minimization (DERM) framework for unsupervised anomaly detection to mitigate the limitation of existing DNN-based methods incurred by the traditional ERM learning principle. 
DERM is well-devised to adaptively control the weights of gradient for corresponding instances via an innovative loss aggregation scheme. Theoretical analysis demonstrates the effectiveness of DERM in suppressing outliers that contaminate the training data. Experimental results show that our DERM framework achieves wide applicability, high flexibility and improved performance on a variety of real-world benchmarks consisting of 18 datasets from diverse domains. In this work, experiments on DERM mainly depend on the assumption that anomalies have larger reconstruction loss. Future studies could investigate the effectiveness of aggregation frameworks on anomaly detectors that adopt different format of loss.

\bibliography{ref} 
\bibliographystyle{splncs04}

\end{document}

%% file: math_commands.tex
\usepackage{amsmath,amsfonts,bm}

\def\eqref#1{equation~\ref{#1}}

\def\1{\bm{1}}

\DeclareMathAlphabet{\mathsfit}{\encodingdefault}{\sfdefault}{m}{sl}
\SetMathAlphabet{\mathsfit}{bold}{\encodingdefault}{\sfdefault}{bx}{n}

